\documentclass[11pt,a4paper,xcolor=table]{article}
\usepackage{times,latexsym}
\usepackage{url}
\usepackage[T1]{fontenc}

\usepackage[acceptedWithA]{tacl2021v1}

\usepackage{xspace,mfirstuc,tabulary}

\newif\iftaclinstructions
\taclinstructionsfalse 
\iftaclinstructions
\renewcommand{\confidential}{}
\renewcommand{\anonsubtext}{(No author info supplied here, for consistency with
TACL-submission anonymization requirements)}
\newcommand{\instr}
\fi

\iftaclpubformat 

\else

\fi

\usepackage{graphicx}
\usepackage{times}
\usepackage{latexsym}

\usepackage{microtype}
\usepackage{footmisc}
\usepackage{scrextend}
\usepackage{array}
\usepackage{adjustbox}
\usepackage{caption}
\usepackage{subcaption}
\usepackage{balance}
\usepackage[T1]{fontenc}

\usepackage{amsmath,breqn}
\usepackage{algorithmic}
\usepackage{array}
\usepackage{txfonts}

\usepackage{mathptmx}
\usepackage{tabularx,booktabs}
\usepackage{longtable}
\usepackage{float}
\usepackage{ragged2e}
\usepackage{xurl}

\title{Improving the Domain Adaptation of Retrieval Augmented Generation (RAG) Models for Open Domain Question Answering}

\author{
  Shamane Siriwardhana
  $^\diamond$, 
  Rivindu Weerasekera$^\diamond$,
  Elliott Wen$^\diamond$, \AND
  Tharindu Kaluarachchi$^\diamond$,
  Rajib Rana$^\dagger$,
  \and
  Suranga Nanayakkara$^\triangle{}^\diamond$
  \\
  $^\diamond$ Augmented Human Lab, Auckland Bioengineering Institute, The University of Auckland \\
  \texttt{firstname@ahlab.org}
  \\
    $^\triangle$ Department of Information Systems \& Analytics, National University of Singapore
  \\
  $^\dagger$ University of Southern Queensland
  \\
  \texttt{Rajib.Rana@usq.edu.au}
}

\begin{document}

\maketitle

\begin{abstract}
Retrieval Augment Generation (RAG) is a recent advancement in Open-Domain Question Answering (ODQA). RAG has only been trained and explored with a Wikipedia-based external knowledge base and is not optimized for use in other specialized domains such as healthcare and news. 
In this paper, we evaluate the impact of joint training of the retriever and generator components of RAG for the task of domain adaptation in ODQA. 
We propose \textit{RAG-end2end}, an extension to RAG, that can adapt to a domain-specific knowledge base by updating all components of the external knowledge base during training. In addition, we introduce an auxiliary training signal to inject more domain-specific knowledge. This auxiliary signal forces \textit{RAG-end2end} to reconstruct a given sentence by accessing the relevant information from the external knowledge base. Our novel contribution is unlike RAG, RAG-end2end does joint training of the retriever and generator for the end QA task and domain adaptation. We evaluate our approach with datasets from three domains: COVID-19, News, and Conversations, and achieve significant performance improvements compared to the original RAG model. Our work has been open-sourced through the Huggingface Transformers library, attesting to our work's credibility and technical consistency.~\footnote{
  This paper is awaiting publication at TACL and this is a pre-MIT Press publication version}

\end{abstract}

\section{Introduction}
Open Domain Question Answering (ODQA)~\cite{lee2019latent,lewis2020question} is an important task in natural language understanding. ODQA methods generally feature a two-stage pipeline: a retriever that selects passages relevant to a given question and a reader that generates the answers from selected passages. Conventionally, these two components are trained separately using ground truth context passages relevant to question-answer (QA) pairs. However, for many real-world scenarios, it is hard to find explicitly annotated context-question-answer triplets ~\cite{lee2019latent,lewis2020retrieval,guu2020realm}. 

Recently, Retrieval Augmented Models (RAGs) have drawn considerable attention from researchers.
RAG consists of a state-of-the-art-neural retriever called Dense Passage Retrieval (DPR)~\cite{karpukhin2020dense} and BART seq2seq language model~\cite{lewis2020pre}. Compared to the conventional two-staged ODQA pipelines, RAG merges the retriever and reader stages into one architecture. Moreover, 
unlike expensive language models with billions of parameters (e.g.,  GPT-3~\cite{brown2020language} and Megatrone-LM~\cite{narayanan2021efficient}) where the model's parametric memory represents the complete knowledge, RAG can also extract knowledge from an external knowledge base. Using both parametric and non-parametric memory generally leads to reduced hallucinations and higher interpretability in tasks like question answering and summarization~\cite{xu2021beyond,komeili2021internet,guu2020realm,lewis2020retrieval}.

In this work, we focus on exploring retrieval augmented architectures for the task of domain-specific open-domain question answering. Although there are several similar retrieval augmented architectures, such as REALM~\cite{guu2020realm} and RETRO~\cite{borgeaud2021improving}, we used Retrieval Augmented Generation (RAG) in our experiments due to its excellent open-source documentation and availability.

When the RAG model is finetuned for downstream QA tasks, the original implementation keeps the encoding of passages and the external knowledge base fixed. This is because re-encoding the external knowledge base is computationally expensive and relies on a sophisticated implementation. Despite not finetuning the passage encodings, the RAG model performs well for datasets with Wikipedia-like knowledge bases because the DPR retriever components have already been trained on Wikipedia-based datasets~\cite{kwiatkowski2019natural,joshi2017triviaqa}. However, the feasibility of adapting RAG to specific ODQA domains such as research papers and news is not well understood. This is a critical research gap to address, as improved domain adaptation can further improve the ODQA performance of RAG.  

This paper explores the feasibility of using RAG in specialized domains for ODQA. In particular, we propose two modifications to the original RAG to improve its domain adaptability.
Motivated by recent end2end retrieval augmented mechanisms~\cite{guu2020realm,sachan2021end,singh2021end}, we first propose a method to finetune the RAG model with its neural retriever and update its knowledge encodings asynchronously during training. We refer to this as \textit{RAG-end2end} since it allows us to update all RAG components during training, including the external knowledge base, the DPR model, and the BART model. Secondly, we propose an auxiliary training signal to help our model learn more domain-specific knowledge. This took the form of generating a concise and factual statement about a document using a self-retrieved set of passages from the provided domain-specific knowledge base. These two modifications offer a unique feature to RAG-end2end over RAG: joint training of the retriever and generator for the end QA task and domain adaptation.
Although asynchronous updates to the knowledge encoder have been proposed before in the REALM, previous work has not evaluated the effects of joint training of the RAG's retriever and the generator for the domain adaptation in ODQA.

We evaluate our proposed approach on three different datasets from three domains: COVID-19 research~\cite{wang2020cord}, Conversations~\cite{wu2021qaconv}, and News~\cite{trischler2016newsqa}. The major finding of our work is that the adaptation of the retriever component plays a critical role in overall domain adaptation performance in RAG-like architectures. Updating only the question encoder without updating the knowledge base encoding could degrade performance. Instead of finetuning the DPR retriever separately, our experiments show that finetuning it as a part of the RAG-end2end mechanism gives better overall results. Our results also show that using the auxiliary signal improves both the retriever component and the overall accuracy. 

In addition, we open-source the implementation of \textit{RAG-end2end} with the HuggingFace Transformers~\cite{wolf2019huggingface} Library\footnote{\href{https://github.com/huggingface/transformers/tree/main/examples/research_projects/rag-end2end-retriever}{Huggingface Transformers implementation}} providing the opportunity for the scientific community to use/test/build on our work.

\section{Background and Related Work}
Open-domain QA systems~\cite{yang2015wikiqa,kwiatkowski2019natural} generally have a two-stage pipeline: passage retrieval (i.e., finding relevant text chunks related to an input question from a knowledge base) and machine comprehension (i.e., generating an answer from a set of selected documents). Traditionally sparse vector methods such as TF-IDF and BM25 are used for document retrieval~\cite{robertson2009probabilistic}. Researchers have recently moved to use dense text representations, which allows modeling textual similarity more semantic level. A recent example  is the `Dense Passage Retriever (DPR)'~\cite{karpukhin2020dense}, which generates embeddings for questions and text passages using two BERT~\cite{devlin2018bert} models. The dot product of the embeddings is used as a similarity score between a question and a passage. DPR has demonstrated that higher retrieval precision results in a higher end-to-end QA accuracy. For the answer generation component of QA systems, recent studies have used either extractive language models like BERT or generative language models like BART/GPT-2~\cite{min2021neurips,lewis2021paq}. 

\subsection{Retrieval Augmented Architecture}
Recently, Retrieval Augmented Architectures ~\cite{lewis2020retrieval,guu2020realm} have drawn a lot of attention due to their explainable, scalable, and adaptable nature. Unlike other open-domain QA architectures, RAG~\cite{lewis2020retrieval} combines the information retrieval stage and answer generation stage in a differentiable manner. It uses a combination of parametric and non-parametric memory, where the parametric memory consists of a pre-trained seq2seq BART~\cite{lewis2019bart} generator, and the non-parametric memory consists of dense vector representations of Wikipedia articles indexed with the FAISS library~\cite{JDH17}. RAG first encodes a question into a dense representation, retrieves the relevant passages from an indexed Wikipedia knowledge base, and then feeds them into the generator. The loss function can finetune both the generator and the question encoder at the same time. Lewis et al. ~\cite{lewis2020retrieval} highlight RAG's ability to perform well in Wikipedia-based general question-answering datasets like Natural Questions~\cite{kwiatkowski2019natural}. Other recent work also highlights how the outputs generated from RAG models are much more factual 
due to RAG being conditioned on the retrieved documents, possibly providing an answer to the hallucination problem of generative language models. Shuster, Kurt, et al.~\cite{shuster2021retrieval} also highlight how RAG reduces hallucinations in knowledge-grounded conversational tasks, where the task is to generate responses to dialogues based on a large Wikipedia knowledge base. \citet{xu2021beyond} illustrate the effectiveness of RAG in chat-bot frameworks and highlight how RAG models are able to recall and summarize conversations compared to standard seq2seq models with only parametric memory. This paper aims to understand how RAG could be extended to an end2end model and adapted to specific domains. To the best of our knowledge, this is the first time RAG is being investigated on domain adaptation for the task of ODQA systems. 

\begin{figure*}[!ht]
\centering
\vspace{-15pt}
\includegraphics[width=1.0\textwidth]{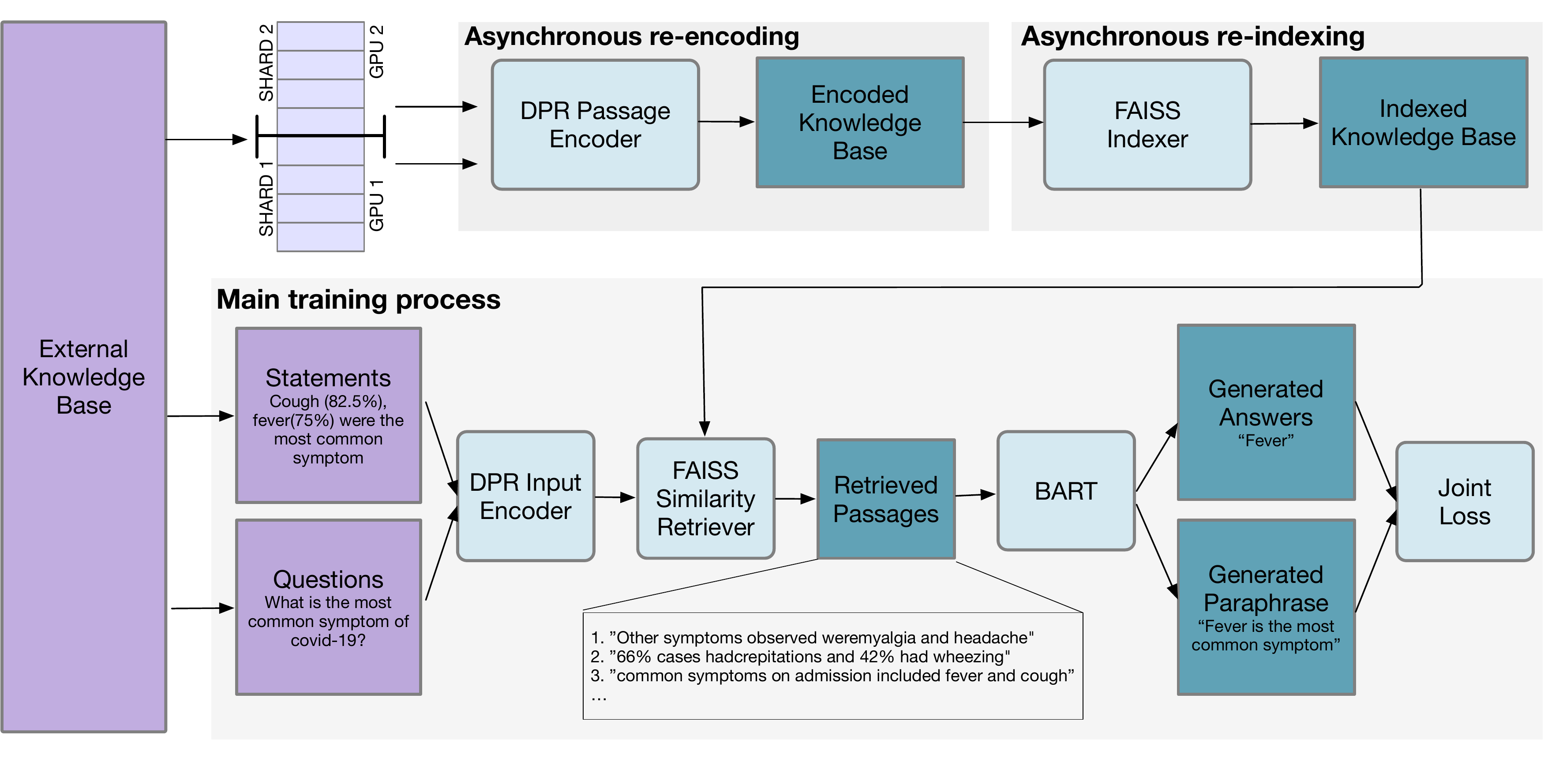}
\caption{System Overview. Our RAG-end2end training architecture uses asynchronous processes to dynamically re-encode and re-index the knowledge base while optimizing a joint QA and paraphrasing signal loss. The training dataset consists of both reconstruction signals and QA pairs. The network learns to generate answers to questions and useful statements jointly. The input to the BART reader is illustrated in Equation~\ref{eq:two_sig}, where the model can differentiate the answer generation task and statement reconstruction task with the use of a control token. During the training, embeddings and the knowledge base index get updated asynchronously.}
\label{fig:overview}
\vspace{-15pt}
\end{figure*}

\subsection{REALM-like end2end Retrieval Augment Architectures}

REALM~\cite{guu2020realm} is a similar Retrieval Augmented model to RAG. REALM introduced a novel masked language pre-training step that involves an end-to-end trainable retriever. In the REALM work, the authors first train the entire model on the masked language prediction task and then fine-tune it on question-answering tasks (keeping the retriever frozen). In comparison to REALM, the original RAG model uses an already trained DPR retriever and conducts partial end-to-end training with a BART reader model. Compared to REALM, RAG is less computationally expensive, and its code is available open-source. We explore and extend the original RAG architecture for domain adaptation in our work. We adapted some concepts of our RAG-end2end extension from REALM. REALM only updates its retriever during the pre-training process that uses the masked language modeling (MLM)~\cite{devlin2018bert} task. Then during the downstream fine-tuning task, REALM keeps its retriever fixed.  
However, the REALM end-to-end training code is not open-sourced, possibly due to its computational complexity. Compared to REALM, RAG is a combination of already pre-trained language models where the users do not need to go through a heavy pre-training stage. Due to these engineering-friendly features and high availability, we conducted our experiments with RAG and extended RAG into an end-to-end trainable retrieval augmentation model. It is also important to highlight that none of the prior work has explored the domain adaptation of retrieval augment models for question answering; instead, most focus on general question answering with Wikipedia-based knowledge bases.

Similar to REALM's end2end architecture, recent work~\cite{sachan2021end} extended RAG and highlighted that the retriever training could improve the overall performance in question-answering datasets like Natural Questions. Compared to our work, the authors did not focus on the domain adaptation of retrieval augment models. The authors mainly explore the ability to train neural retrievers in an end-to-end way using retrieval augment models. Similarly, another related work~\cite{singh2021end} extended retrieval augmented architectures to an end-to-end model and illustrated that it could improve the question answering accuracy. \citet{singh2021end} mainly focused on improving the document reading ability and answer generation rather than domain adaptation.

\section{Model Architecture and Training Procedure}
 In this work, we extend RAG to finetune all components, including the DPR retriever, and dynamically update the external knowledge base during training. We hypothesize that the use of asynchronous updates helps with domain adaptation. Figure~\ref{fig:overview} demonstrates the main workflow of our model. In the following sections, we describe our extensions and training signals.   
\subsection{RAG Retriever and Generator}
The retriever is a DPR~\cite{karpukhin2020dense} model pre-trained on Wikipedia-based question-answering datasets~\cite{kwiatkowski2019natural,joshi2017triviaqa}. It consists of two tower BERT-based networks: the Question Encoder ($E_{Q}$) and the Passage Encoder ($E_{P}$). 
We use their CLS token embeddings as representations for questions and passages. 
The similarity between a question ($q$) and a passage ($p$) is calculated by taking the dot product of the two embeddings as shown in Equation~\ref{eq:dpr}.
\begin{equation}
sim(p,q) \propto E_{Q}(q)^{T}E_{P}(p).
\label{eq:dpr}
\vspace{-10pt}
\end{equation}

\vspace{1pt}
RAG's generator consists of a pre-trained BART~\cite{lewis2019bart} seq2seq language model. To train these retriever and generator components, RAG enhances the traditional sequence-to-sequence cross-entropy loss function by setting the retrieved passages as a latent variable ($Z$)~\cite{guu2020realm,lewis2020retrieval}. The loss value of generating each token is marginalized on the probability of selecting documents given a context $X$ (i.e., Document Score $p(Z|X)$). The formula (RAG-Token-Loss) can be written as illustrated in Equation~\ref{eq:rag-con}.
\vspace{-15pt}
\begin{multline}
P_{RAG-Token-Loss}(y|x) =\\ \prod_{i}^{n}\sum_{z \varepsilon_{top-k} P(.|x)}P_{\eta}(z|x)P_{\theta}(y_{i}|x,z,y_{1:i-1})
\label{eq:rag-con}
\end{multline}

\subsection{Indexing of the External Knowledge Base}
Before the training phase, we need to encode all passages in the external knowledge base using $E_{P}$. Then we need to retrieve similar passages from the external knowledge base given the output from $E_{Q}$.  This process mainly involves dot product calculation between input question embeddings and encoded passages. The retrieval process will likely result in a performance bottleneck during the training since there are usually millions of passages in the knowledge base. To address this issue, RAG adopts the FAISS indexing approach proposed in ~\cite{JDH17}. With the help of the indexes, we can skip a considerable amount of repeated computation and significantly accelerate the retrieval process. 

\subsection{End-to-End Retriever Training}
Although the DPR module makes use of two BERT models ($E_{P}$,$E_{q}$), the original RAG architecture only fine-tunes the question encoder $E_{Q}$ in the retriever. The passage encoder $E_{P}$ and the external knowledge base's encoding are fixed during the training phase. In other words, the pre-trained passage encoder of DPR is only used once to encode the external knowledge base.
The RAG authors suggest that such a design performs well for Wikipedia-based ODQA datasets~\cite{kwiatkowski2019natural,joshi2017triviaqa}. Such settings work because the DPR model was also pre-trained with Wikipedia-based datasets, and their experiment uses an external knowledge base consisting of Wikipedia articles.

However, it may be beneficial to fine-tune all the DPR components during RAG training for domain adaptation since the model needs access to different domain-specific external knowledge bases. 
In this work, we introduce \textit{RAG-end2end}, where we augment RAG to be fully end-to-end trainable. We fine-tune the passage encoder and question encoder and then update the index of the external knowledge base during the training process.

It is straightforward to propagate gradients to both the passage and question encoders with RAG's loss function. Because this loss function employs the passage selection probability known as doc-score($p_{\eta}(z|x)$ term illustrated in Equation~\ref{eq:rag-con}).
However, for it to have a true effect on the overall model training process, we have to iteratively update the embeddings with the updated context encoder and then update the index of the external knowledge base. In other words, we need to re-encode and re-index the knowledge base using the updated passage encoder. 
When the external knowledge base possesses tens of millions of passages, the re-encoding and re-indexing steps can be very time-consuming. Re-encoding can take several GPU hours, and re-indexing with FAISS can take several CPU hours, depending on the size of the knowledge base. 
Therefore, it is inefficient to stall the training loop while the re-encoding re-indexing steps are being carried out.

To have an efficient training mechanism, we designed our training framework into three main processes: (1) The main training loop, which updates the gradients, (2) Re-encoding processes with several GPUs that update the knowledge-base encoding with the updated DPR's context encoder, and (3) A Re-indexing process that uses FAISS to build an index with the updated encoding. Figure~\ref{fig:overview} illustrates these three processes. Our implementation uses two asynchronous processes to re-encode and re-index the external knowledge base that runs independently to the main training loop. We first distribute the external knowledge base to a set of GPUs that are not used in the main training loop. Then we encode the passages with an updated passage encoder which we call the re-encoding process. Once the re-encoding process has finished, we re-index the knowledge base in another parallel process that uses FAISS (re-indexing process). Inside the main training loop, we ensure that the re-indexing process always starts after finishing the re-encoding process. Then as soon as the new index of the external knowledge base is created, we load that to the main training loop. Once the new index loading is completed again, we start the re-encoding process, which repeats the entire embedding updating process.
It is important to note that the first re-encoding process should get finished, and new embeddings should get saved to the hard disk before the start of the FAISS indexing process. If the knowledge base is not entirely updated with the new embeddings, the re-indexing process fails. We use python multiprocessing handles to keep the order, and re-indexing and re-encoding processes are only asynchronous with respect to the main training loop process.  
The number of steps between each re-encoding process depends on the size of the dataset. To test the number of steps between the knowledge-base updates, we experimented with a knowledge base consisting of 250,000 passages and used four dedicated GPUs for the re-encoding process with a batch size of 32 each. Our computation machine consists of 96 CPU cores. We found that it takes an average of 750 updates. However, the computation time can be easily improved when using more GPUs for encoding and using a machine with a higher number of CPU cores (FAISS indexing process depends on the number of CPU cores). These steps are repeated throughout the training loop. Since the training and knowledge base's index update processes are running asynchronously, it may result in stale gradients. This, however, does not significantly degrade the model performance according to previous research~\cite{guu2020realm}.

\subsection{Statement Reconstruction}\label{sec:recon}
We explore the incorporation of statement reconstruction as an auxiliary signal assuming that it forces the model to gain more domain-specific knowledge. As illustrated in  Figure~\ref{fig:overview}, we first encode input statements using the input/question encoder ($E_{Q}$). Then the retriever retrieves the most similar set of passages from the indexed external knowledge base by conducting a similarity search. Afterward, the final output generator attempts to re-construct the input statements using only the selected support set of documents. We ensure that the external knowledge base does not contain the input statement to prevent the model from overfitting on just the lexical content. To differentiate the paraphrasing signal from the QA signal, we prepend a special token $<p>$ (represents passages) in front of the reconstruction statements, which acts as a control token in the seq2seq language modeling~\cite{raffel2019exploring,keskar2019ctrl}. Concretely, when training the RAG architecture on QA pairs, the questions are prepended to the retrieved passages before being fed to the BART generator. As illustrated in Equation~\ref{eq:two_sig}, for the input reconstruction signal, we only prepend the $<p>$ token to the retrieved passages before feeding them to the BART generator.

\vspace{-20pt}
\begin{equation}
\label{eq:two_sig}
\resizebox{.9 \columnwidth}{!}{
$\begin{matrix}
\textsc{QA Input: } & \left \langle \texttt{Question} \right \rangle + \left \langle \texttt{Retrieved Passages}  \right \rangle 
\\ 
\textsc{Reconstruction Input: } & \left \langle \texttt{<p>} \right \rangle + \left \langle \texttt{Retrieved Passages}  \right \rangle 
\end{matrix}
$}
\end{equation}
\vspace{-25pt}

\section{Experiments \& Results} 
\subsection{Domain Specific Dataset Setup}
In this work, our main intention is to explore the adaptation of domain-specific retrieval augmentation with regard to ODQA. As mentioned in the recent work~\cite{lewis2020retrieval}, most ODQA datasets like Natural Questions~\cite{kwiatkowski2019natural} , TriviaQA~\cite{joshi2017triviaqa}, WebQuestions~\cite{berant2013semantic}, and CuratedTrec~\cite{baudivs2015modeling} are answered with Wikipedia-based knowledge-bases. Since neural retrievers like DPR are already trained with Wikipedia-based datasets, it is hard for us to explore the domain adaptation of RAG fairly in this setting. Therefore,  we selected three domain-specific datasets for our experiment: COVID-19 QA, News QA, and Conversation QA. Since the availability of domain-specific ODQA datasets is minimal, in our work, we open-source all domain-specific knowledge-bases and question-answer pairs to support future research\footnote{\href{https://drive.google.com/drive/folders/1up3yKcJFArBQ6e0F_6n_mfW1VPHxA20A?usp=sharing}{domain specific datasets}}.

\noindent\textbf{COVID-19 QA Domain}\\
\noindent\emph{Knowledge Base Generation:}
To create the external knowledge base, we use 5,000 full-text scientific articles extracted from the CORD-19~\cite{wang2020cord} dataset. The external knowledge base is created with 250,000 100-word passages. Each passage is pre-pended with the title of the research paper.

\noindent\emph{Reconstruction Statement Generation:}
We use sentences from the abstract section of research articles for the reconstruction signal. We first extract the abstract sections in 10K papers and split them into sentences using the NLTK library~\cite{loper2002nltk}. 
We filter out the sentences that are too short (less than 15 words) or too long (more than 35 words). In this process, approximately 50,000 abstract statements are generated. It is important to note that when generating the knowledge base, we exclude the abstract sections. 

\noindent\emph{Synthetic QA Generation:}\label{sec:coviddata} 
In this domain, we only use synthetic data for training and validation.
Following the prior work~\cite{shakeri2020end}, we use a BART seq2seq model trained on the SQuAD dataset  ~\cite{rajpurkar2016squad} to generate synthetic QA pairs given a passage. We used the Squad dataset's passages as the input and corresponding question-answer pairs as the expected output. We trained a BART-large checkpoint for two epochs. Then, we followed round-trip consistency~\cite{alberti2019synthetic} to filter synthetic QA pairs. Our final synthesized QA dataset consisted of 225,000 QA pairs. We use 90\% of these QA pairs as training data and 10\% as validation data. As the test data, we use 2000 human-labeled question-answer pairs from the COVID-QA dataset~\cite{moller-etal-2020-covid}.

\noindent\textbf{News QA Domain}\\
\noindent\emph{Knowledge Base Generation:}
We extract 85,000 100-word passages as the knowledge base using 10,000 news articles from the NewsQA dataset~\cite{trischler2016newsqa}. 

\noindent\emph{Reconstruction Statement Generation:} 
We extract corresponding news summary sentences from the CNN/DM dataset~\cite{hermann2015teaching} for the reconstruction signal. Every article consists of more than one summary sentence. However,  we use the first sentence as the title of the article, which we used in knowledge base generation and the rest of the statements as reconstruction statements. 
Our final dataset contains 35,000 summary statements.   

\noindent\emph{QA Generation:}
The NewsQA dataset~\cite{trischler2016newsqa} consists of 100,000 human annotated QA pairs from 10,000 news articles from the CNN/DM dataset~\cite{hermann2015teaching}. We use the train (90,000), valid (5,000) and test (5,000) splits given in the dataset to train and evaluate our model. All questions in the NewsQA dataset focus on the high-level content of articles. So,  to answer these questions, the model must access a large span of passages to conduct the reasoning process.

\noindent\textbf{Conversation QA Domain}\\ 
\noindent\emph{Knowledge Base Generation:}
We create the external knowledge base of 110,000 passages by splitting the 10,000 conversations given in the QAConv dataset~\cite{wu2021qaconv} into passages, each with at most 100 words. We prepend the identifier of each conversation (found in the original dataset) as the title of the passages. We also appended the speaker's name, followed by the ":" symbol, to the starting position of each dialogue to keep each conversation connected to its speakers.

\noindent\emph{Reconstruction Statement Generation:} 
We use the state-of-the-art abstractive conversation summarization model\footnote{\href{https://github.com/salesforce/ConvSumm/tree/master/CODS\#load-trained-model}{salseforce checkpoint}}~\cite{wu2021controllable} to generate one-sentence (TLDR) summary (approximately 45 words per conversation). We then use this as the auxiliary signal. We only generate summaries of conversations with more than 45 words. By doing this, we collect 35,000 synthetic summary/reconstruction statements. 

\noindent\emph{QA Generation:} 
We use the QAConv dataset~\cite{wu2021qaconv}, which contains 35,000 QA pairs generated from 10,000 conversations that involved two or more parties. We use the train (25,000), valid (5,000) and test (5,000) splits given in the dataset to train and evaluate our model.

\subsection{Training and Evaluation Setup}\label{sec:eval}
We use the HuggingFace-Transformers~\cite{wolf2019huggingface} library to implement the RAG-end2end architecture. We initialize the DPR and BART models of using the open-source HuggingFace checkpoints\footnote{\href{https://huggingface.co/facebook/rag-token-base}{rag-token-base checkpoint}}. Prior to fine-tuning, we index and encode the external knowledge base using FAISS.
We select HNSW FLAT as the indexing mechanism (with 128 bi-directional links). We use 100 words as the maximum passage length as suggested by the prior RAG work~\cite{lewis2020pre}. During training, we use six Tesla V100 GPUs with 32 GBs of memory. Four of them are used for training, and two are used for re-encoding. We train each RAG model variant~\ref{sec:mod} for ten epochs and select the final checkpoint with the highest validation accuracy.

We use the Exact Match (EM), F1 score, and Top-K retrieval accuracy as evaluation metrics. The EM score computes the word level exact match between the predicted answer and the real answer. The F1-score calculates the number of words in the predicted answer that are aligned with the real answer regardless of the order. The Top-k retrieval accuracy is calculated by matching the answer strings with the contents of the retrieved k passages.

We compare RAG and RAG-end2end in the following five scenarios.
\begin{enumerate}\label{sec:mod}
    \item \textbf{RAG-original}. This model is
    finetuned on the natural question dataset~\cite{kwiatkowski2019natural} with the Wikipedia knowledge base and serves as the non-domain adapted baseline~\footnote{\href{https://huggingface.co/facebook/rag-token-nq}{public rag-token-nq checkpoint}}. This model is not finetuned with domain-specific question-answer pairs, and we report the zero-shot performance.
    \item \textbf{RAG-original-QA}. This is the original RAG model finetuned with only domain-specific question-answer pairs.
     \item \textbf{RAG-end2end-QA}. This is the RAG model with our end2end retriever extensions and finetuned only with domain-specific question-answer pairs.
    \item \textbf{RAG-original-QA $+$ R}.  This is the RAG original model finetuned with both domain-specific question-answer pairs and our reconstruction signal.
   \item \textbf{RAG-end2end-QA $+$ R}. This is the RAG model with our end2end retriever extensions and trained with both question-answer pairs and our reconstruction signal.
\end{enumerate}
We present the results of each scenario in Table~\ref{tab:main}.

\subsection{Effect of End-to-End Retriever Training on Domain Adaptation}\label{sec:effect_end2end}
We first test if finetuning of both the passage encoder and question encoder of the RAG's retriever while updating the external knowledge base would improve domain adaptation. We compare the performance of  RAG-original-QA and RAG-end2end-QA, isolating any performance improvement due to the reconstruction signal. The results in Table~\ref{tab:main} illustrate that RAG-end2end-QA significantly outperforms RAG-original-QA on all metrics -- EM, F1, Top-5, and Top-20 -- across all three domains. The improvements in the EM score varied from 1.13 points in the News domain to 12.16 points in the Conversation domain.

Evaluating the performance of passage retrieval using Top-5 and Top-20 scores, we see a marked increase of around 25 points in the conversation domain, with the other domains showing improvements of between 4.7 to 6.6 points. 

Above all, these results suggest that fine-tuning both the passage and question encoders of the RAG's retriever while updating the external knowledge base can improve domain adaptation.

\begin{table}[]

\def\arraystretch{1.3}
\resizebox{\columnwidth}{!}{%
\begin{tabular}{lllll}
\multicolumn{1}{c|}{Model Name}& \multicolumn{1}{c|}{EM}    & \multicolumn{1}{c|}{F1}    & \multicolumn{1}{c|}{Top-5} & \multicolumn{1}{c}{Top-20} \\ \hline
\multicolumn{5}{c}{COVID-19 Domain}                                                                                                                                                                \\ \hline
\multicolumn{1}{l|}{(1) RAG-original}                                          & \multicolumn{1}{l|}{0.0}  & \multicolumn{1}{l|}{4.73}  & \multicolumn{1}{l|}{10.56}  & \multicolumn{1}{l}{15.69}   \\
\multicolumn{1}{l|}{(2) RAG-original-QA}                                        & \multicolumn{1}{l|}{2.95}  & \multicolumn{1}{l|}{12.01} & \multicolumn{1}{l|}{12.29}  & \multicolumn{1}{l}{18.43}  \\
\multicolumn{1}{l|}{(3) RAG-end2end-QA}                                        & \multicolumn{1}{l|}{8.08}  & \multicolumn{1}{l|}{ 18.38} & \multicolumn{1}{l|}{ 19.85} & \multicolumn{1}{l}{26.91}  \\
\multicolumn{1}{l|}{(4) RAG-original-QA+R}                                      & \multicolumn{1}{l|}{3.66}  & \multicolumn{1}{l|}{12.20} & \multicolumn{1}{l|}{12.79} & \multicolumn{1}{l}{18.45}  \\
\multicolumn{1}{l|}{(5) RAG-end2end-QA+R}                                      & \multicolumn{1}{l|}{8.32}  & \multicolumn{1}{l|}{19.57} & \multicolumn{1}{l|}{23.05} & \multicolumn{1}{l}{31.23}  \\ \hline

\multicolumn{5}{c}{News Domain}                                                                                                                                                                    \\ \hline

\multicolumn{1}{l|}{(1) RAG-original}                                          & \multicolumn{1}{l|}{4.33}  & \multicolumn{1}{l|}{7.92}  & \multicolumn{1}{l|}{19.46} & \multicolumn{1}{l}{30.33}  \\
\multicolumn{1}{l|}{(2) RAG-original-QA}                                        & \multicolumn{1}{l|}{7.26}  & \multicolumn{1}{l|}{14.26} & \multicolumn{1}{l|}{22.86} & \multicolumn{1}{l}{34.55}  \\
\multicolumn{1}{l|}{(3) RAG-end2end-QA}                                        & \multicolumn{1}{l|}{8.39}  & \multicolumn{1}{l|}{16.31} & \multicolumn{1}{l|}{28.89} & \multicolumn{1}{l}{41.20}   \\
\multicolumn{1}{l|}{(4) RAG-original-QA+R}                                      & \multicolumn{1}{l|}{8.62}  & \multicolumn{1}{l|}{16.46} & \multicolumn{1}{l|}{27.28} & \multicolumn{1}{l}{39.56}  \\
\multicolumn{1}{l|}{(5) RAG-end2end-QA+R}                                      & \multicolumn{1}{l|}{14.08} & \multicolumn{1}{l|}{23.7}  & \multicolumn{1}{l|}{39.67} & \multicolumn{1}{l}{50.95}  \\ \hline

\multicolumn{5}{c}{Conversation Domain}                                                                                                                                                            \\ \hline

\multicolumn{1}{l|}{(1) RAG-original}                                          & \multicolumn{1}{l|}{5.49}  & \multicolumn{1}{l|}{9.27}  & \multicolumn{1}{l|}{12.14} & 20.02                       \\
\multicolumn{1}{l|}{(2) RAG-original-QA}                                        & \multicolumn{1}{l|}{12.09} & \multicolumn{1}{l|}{20.05} & \multicolumn{1}{l|}{22.73} & 32.05                       \\

\multicolumn{1}{l|}{(3) RAG-end2end-QA}                                        & \multicolumn{1}{l|}{24.25} & \multicolumn{1}{l|}{36.05} & \multicolumn{1}{l|}{46.01} & 55.55  
\\
\multicolumn{1}{l|}{(4) RAG-original-QA+R}                                      & \multicolumn{1}{l|}{14.21} & \multicolumn{1}{l|}{24.62} & \multicolumn{1}{l|}{26.32} & 36.21                       \\
\multicolumn{1}{l|}{(5) RAG-end2end-QA+R}
 & \multicolumn{1}{l|}{25.95} & \multicolumn{1}{l|}{37.96} & \multicolumn{1}{l|}{49.11} & 58.75                       \\ \hline

\end{tabular} 
}%
\caption{Domain adaptation Performance of different RAG models used in our experiments. We illustrate the results related to all three domains. Details about each model are described in Section~\ref{sec:eval}}
\label{tab:main}
\vspace{-10pt}
\end{table}

\subsection{Effect of Adding the Statement-Reconstruction Auxiliary Task}\label{sec:effect_auxilary}

\begin{table*}[h!]
\centering
\def\arraystretch{3}
\resizebox{0.95\textwidth}{!}{%
\begin{tabular}{p{0.12\textwidth}|p{0.25\textwidth}|p{0.4\textwidth}|p{0.25\textwidth}}
Domain &
  Input Statement &
  Most Similar Retrieved Document &
  Reconstructed Statement \\ \hline
COVID-19 &
  \textless{}p\textgreater Cough (82.5\%), fever (75\%), and malaise (58.8\%) were the most common symptoms, and crepitations (60\%), and wheezes (40\%) were the most common signs. &
  Most common signs and symptoms on admission included fever and cough . Of all children, 32\% had complaint of difficulty in respiration. Other symptoms observed were myalgia, headache and vomiting . On examination, 66\% cases had crepitations and 42\% had wheezing. Hypoxemia was observed in 31\% cases at admission &
  The most common signs and symptoms on admission were fever and cough, and 32\% had complaint of difficulty breathing \\ \hline
News &
  \textless{}p\textgreater Capsule was carrying South Korea's first astronaut . &
  MOSCOW, Russia Russian space officials say the crew of the Soyuz space ship is resting after a rough ride back to Earth. A South Korean bioengineer was one of three people on board the Soyuz capsule. The craft carrying South Korea's first astronaut landed in northern Kazakhstan on Saturday, 260 miles off its mark, they said. &
  Soyuz capsule carrying South Korea's first astronaut lands 260 miles off its mark. \\ \hline
Conversation &
  \textless{}p\textgreater The Supreme Court will hear the case on the grounds of First Amendment protection of free speech. &
  (PETITIONER): Yes, Your Honor. Mr. Tory, who was appearing pro se in the trial court, from the very outset objected that he was being held liable for speech protected by the First Amendment. \textless{}end\textgreater \$ &
  Justice Souter and MR. CHEMERINSKY are arguing that the injunction in the case of Tory should not be applied \\ \hline
\end{tabular}%

}

\caption{Examples of Reconstructed Statements. Reconstructions generally capture the context of the retrieved documents and are similar to the input statement but are not always factually 100\% correct (e.g. COVID-19 example). Input statement column shows the input to the model with the special $<$p$>$ token. The Retrieved Documents shows a snap-shot of the top-retrieved document used to re-construct the statement}
\label{tab:recons}
\vspace{-10pt}
\end{table*}
In this experiment, we test our next hypothesis: adding the auxiliary training signal of statement reconstruction along with QA pairs improves domain adaptation. We compare the performance of RAG-end2end with and without the reconstruction signal by comparing the performance of RAG-end2end-QA $+$ R and RAG-end2end-QA in Table~\ref{tab:main}. This shows that RAG-end2end-QA $+$ R outperforms RAG-end2end-QA for all three domains. The range of increases in the EM scores varied from 1.7 points in the conversation domain to an  8.39 points increase in the News domain.
The top-20 retrieval accuracy also increased in a range between 3.2 to 8 points.

We further compare the effect of adding the reconstruction signal to RAG-original by comparing RAG-Original-QA with RAG-Original-QA $+$ R. We find that even without the end2end extension, the reconstruction signal improves the performance moderately. This improvement in the EM score ranged from 0.84 points in the COVID-19 domain and 3.12 points in the Conversation domain.

Finally, we highlight the overall improvement of our contributions by comparing RAG-Original-QA with RAG-end2end-QA$+$ R. As the most significant improvement; we highlight the 13-point improvement of EM score for the Conversation domain. For retrieval performance, we highlight the 27-point improvement in the top 5 and 16 improvements in the top 20 for the Conversation domain.  

To demonstrate the reconstruction statement generation, we provide an example of the generated reconstruction output of given a statement for each domain using the  RAG-end2end-QAR $+$ R model in Table~\ref{tab:recons}. The second column contains the input statements with the special token $<p>$, the third column shows a snapshot of retrieved top-5 documents, and the final column shows the re-constructed statements. As the reconstruction statements demonstrate, we highlight that the model can generate statements close enough to the input.

\subsection{Retriever's domain adaptation with RAG-end2end}\label{ex:dpr}

An important part of our RAG-end2end extension is updating the entire DPR retriever during training. 
Previous work~\cite{ma2020zero} has explored the importance of the domain adaptation of neural retrievers and highlighted the performance gains in domain-specific retrieval tasks. 
We argue, based on our above-mentioned RAG end2end's retriever performances and prior work, that when adapting RAG to various domains, having a domain-specific retriever plays a key role in achieving good performance. However, this end-to-end RAG finetuning can get computationally costly, especially with the number of passages in the external knowledge base where they should get re-encoded and re-indexed. Instead of finetuning DPR as a part of RAG-end2end, an alternative approach is to finetune DPR on domain-specific data separately on its vector similarity-based loss function~\cite{karpukhin2020dense} and then initializing the RAG architecture prior to finetuning with the QA data. 
We explore if RAG-end2end can perform on par if we initialize a RAG model with an independent domain-adopted DPR model. 
This helps us further understand the ability of the RAG-end2end extension to finetune the retriever with domain-specific data.\\
\\

\noindent\textbf{Standalone DPR fine tuning with domain specific data}\\
 
The standalone DPR can be finetuned if we have access to gold-standard passages that contain the answers for given questions and hard negative passages which consist of similar details to the question but not the exact answers. 
DPR uses a dot-product-based similarity loss, capturing the similarity between the correct passage for the question while comparing with some hard-negative examples~\cite{karpukhin2020dense} (which are lexically similar but do not contain the answer)~\cite{karpukhin2020dense}.  
We use the deep-haystack framework\footnote{\href{https://github.com/deepset-ai/haystack}{deepset-ai}} to finetune DPR for each domain using domain-specific data. We created finetuning datasets for all three domains. First, for the Covid-19 domain, we utilized the synthetic question-answer pairs and their relevant passages that consist of 100 words. The use of domain-specific synthetic QA pairs for DPR finetuning has already shown permanence improvements~\cite{ma2020zero}. For hard-negative examples, we used BM-25 lexical matching search as mentioned by the DPR authors, where we retrieved passages that do not contain the answer based on their lexical similarity with the question. Although for the News domain and the Conversation domain, we have a supervised dataset where we can map questions into the correct passage, we did not get better results after finetuning the original DPR using the supervised data. The main reason for the degradation of performance is the length of the correct passage related to the question. In both News and Conversation domains, most of the questions come from longer passages, whereas the pre-trained DPR only accepts 100-word passages. To mitigate this issue, we generated synthetic question-answer pairs with the external knowledge bases of news and Conversation domains similar to the COVID-19 domains by following the same procedure mentioned in Section~\ref{sec:coviddata}. Then the hard-negative examples were also mined according to the above-mentioned BM-25 lexical matching method. After training, we evaluate the DPR retrieval accuracy using the test dataset and external knowledge base for each domain, similar to the RAG's retrieval evaluation we conducted in Section~\ref{sec:eval}

\begin{table}[]
\centering

\def\arraystretch{1.3}
\resizebox{0.8\columnwidth}{!}{%
\begin{tabular}{lll}
\multicolumn{1}{c|}{Model Name}& \multicolumn{1}{c|}{Top-5} & \multicolumn{1}{c}{Top-20} \\ \hline
\multicolumn{3}{c}{COVID-19 Domain}                                                                                                                                                                \\ \hline
\multicolumn{1}{l|}{(1) DPR-original}                                           & \multicolumn{1}{l|}{9.39}  & \multicolumn{1}{l}{14.72}   \\

\multicolumn{1}{l|}{(2) DPR-domain-adapted}                                         & \multicolumn{1}{l|}{13.66}  & \multicolumn{1}{l}{20.01}  \\

\multicolumn{1}{l|}{(3) DPR-RAG-end2end}                                         & \multicolumn{1}{l|}{ 20.64} & \multicolumn{1}{l}{28.21}  \\
 
\hline

\multicolumn{3}{c}{News Domain}                                                                                                        \\ 
\hline

\multicolumn{1}{l|}{(1) DPR-original}                                         & \multicolumn{1}{l|}{20.95} & \multicolumn{1}{l}{31.04}  \\

\multicolumn{1}{l|}{(2) DPR-domain-adapted}                                         & \multicolumn{1}{l|}{20.98} & \multicolumn{1}{l}{31.92}  \\

\multicolumn{1}{l|}{(3) DPR-RAG-end2end}                                         & \multicolumn{1}{l|}{39.67} & \multicolumn{1}{l}{50.95}   \\
 \hline

\multicolumn{3}{c}{Coversation Domain}                                                                                                                                                            \\ \hline

\multicolumn{1}{l|}{(1)  DPR-original}                                          & \multicolumn{1}{l|}{15.15} & 23.95                     \\

\multicolumn{1}{l|}{(2) DPR-domain-adapted}                                      & \multicolumn{1}{l|}{23.15} & 34.53                       \\

\multicolumn{1}{l|}{(3) DPR-RAG-end2end}                                        & \multicolumn{1}{l|}{49.11} & 58.75  
\\
\hline
\end{tabular}%
}
\caption{Comparison of DPR models finetunned on domain specific data against publicly available DPR checkpoint which is trained on Wikipedia domain for all three domains.}
\label{tab:domain-dpr}
\vspace{-10pt}
\end{table}

Table~\ref{tab:domain-dpr} compares \textbf{(1) DPR-orignal}, which is the publicly available checkpoint\interfootnotelinepenalty=10000 \footnote{\href{https://github.com/facebookresearch/DPR\#dense-passage-retrieval}{DPR-checkpoint} } trained on Wikipedia, with \textbf{(2) DPR-domain-adapted}, which is the finetuned model with DPR's original loss function.  
The \textbf{(3) DPR-RAG-end2end} is  the retrieval part of RAG-end2end-QA $+$ R from Table~\ref{tab:main} for comparison. We include the DPR-RAG-end2end model to highlight the improvement of the DPR model as a result of RAG-end2end training with both training signals. When comparing the DPR-RAG-end2end model with the other variants in Table~\ref{tab:main}, we observe that the RAG-end2end architecture significantly improves the DPR's domain adaptation for all three domains. Therefore, in future work, RAG-end2end could be used as a way to train a neural retriever, which could benefit even for retrieval-only applications.

As shown in Table~\ref{tab:domain-dpr}, we observe that fine-tuning DPR models on the original DPR loss function using domain-specific data improves the overall retrieval performance for each domain. For the Covid-19 and Conversation domains, there's a clear improvement in the top-5 and top-20 retrieval accuracies. We observed almost the same results for the News domain compared to the original DPR. This could be due to similar kinds of data in Wikipedia, which were originally used to train the DPR and CNN/DM text.

Overall as illustrated in Table~\ref{tab:domain-dpr}, we highlight the fact that the RAG-end2end's loss function has the ability to adapt the DPR to specific domains better than fine-tuning the DPR with the passages and question pairs for each domain. The improvements for all three domains in top-5 and top-20 retrieval accuracies of  DPR-RAG-end2end  compared to  DPR-original and DPR-domain-adapted is noticeable. These results further highlight the ability of RAG-end2end to fine-tune or improve its retriever.
\vspace{.2cm}

\noindent\textbf{Initializing RAG with domain adapted DPR prior to finetuning}\\
Next, we investigate the performance of RAG models when initialized with a domain-adapted DPR.
We initialize RAG's question encoder and the passage encoder with DPR-domain-adapted (from trained models illustrated in Table~\ref{tab:domain-dpr}) and finetune RAG with the settings of  RAG-original-QA$+$R. The objective is to compare how the RAG models initialized with domain adopted DPR models perform in comparison to using the RAG-end2end extension.

\begin{table}[h!]
\centering

\def\arraystretch{1.3}
\resizebox{\columnwidth}{!}{%
\begin{tabular}{lllll}
\multicolumn{1}{l|}{Model Name}                         & \multicolumn{1}{l|}{EM score} & \multicolumn{1}{l|}{F1 Score} & \multicolumn{1}{l|}{Top-5} & Top-20 \\ \hline
\multicolumn{5}{c}{COVID-19 Domain} \\ \hline

\multicolumn{1}{l|}{ (1) RAG-original-QA+R}   & \multicolumn{1}{l|}{3.66}     & \multicolumn{1}{l|}{12.12}    & \multicolumn{1}{l|}{12.79}  & 18.45  \\ 

\multicolumn{1}{l|}{ (2) RAG-original-QA+R (DPR-adapted)}   & \multicolumn{1}{l|}{7.36}     & \multicolumn{1}{l|}{17.91}    & \multicolumn{1}{l|}{22.39}  & 30.87  \\ 

\multicolumn{1}{l|}{ (3) RAG-end2end-QA+R}   & \multicolumn{1}{l|}{8.32}     & \multicolumn{1}{l|}{19.51}    & \multicolumn{1}{l|}{23.05}  & 31.23  \\  \hline

\multicolumn{5}{c}{News Domain} \\ \hline

\multicolumn{1}{l|}{ (1) RAG-original-QA+R}   & \multicolumn{1}{l|}{8.62}     & \multicolumn{1}{l|}{16.46}    & \multicolumn{1}{l|}{27.28}  & 39.56  \\ 

\multicolumn{1}{l|}{ (2) RAG-original-QA+R (DPR-adapted)}   & \multicolumn{1}{l|}{10.92}     & \multicolumn{1}{l|}{19.44}    & \multicolumn{1}{l|}{30.72}  & 41.9  \\ 

\multicolumn{1}{l|}{ (3) RAG-end2end-QA+R}   & \multicolumn{1}{l|}{14.08}     & \multicolumn{1}{l|}{23.7}    & \multicolumn{1}{l|}{39.67}  & 50.95  \\

 \hline
\multicolumn{5}{c}{Conversation Domain} \\ \hline

\multicolumn{1}{l|}{ (1) RAG-original-QA+R}   & \multicolumn{1}{l|}{14.21}     & \multicolumn{1}{l|}{24.62}    & \multicolumn{1}{l|}{26.32}  & 36.21  \\

\multicolumn{1}{l|}{ (2) RAG-original-QA+R (DPR-adapted)}   & \multicolumn{1}{l|}{15.78}     & \multicolumn{1}{l|}{25.47}    & \multicolumn{1}{l|}{29.01}  & 40.03  \\ 

\multicolumn{1}{l|}{ (3) RAG-end2end-QA+R}   & \multicolumn{1}{l|}{25.95}     & \multicolumn{1}{l|}{37.96}    & \multicolumn{1}{l|}{49.11}  & 58.75  \\ 

\hline
\end{tabular}%
}
\caption{Comparing the effect of RAG-end2end extension, against initializing RAG-original models with domain adapted DPR models prior to the fine-tuning (Please check the Table~\ref{tab:main}). We use the independently domain adapted DPR models illustrated in Table~\ref{tab:domain-dpr}
}
\label{tab:dpr_init}
\end{table}

Table~\ref{tab:dpr_init} demonstrates results from four models. (1) RAG-original-QA$+$R and (3) RAG-end2end-QA$+$R are taken from the main results (Table~\ref{tab:main}). The \textbf{(2) RAG-original-QA$+$R (DPR-adapted)} model was first initialized with a domain-adopted DPR model (from Table~\ref{tab:domain-dpr}) before being finetuned with domain-specific QA pairs and re-construction signals with the RAG-original settings. 

The results in Table~\ref{tab:dpr_init} indicate that for all domains, finetuning the RAG-original with a domain-adapted DPR gives higher performance than finetuning the RAG-original with the usual DPR model checkpoint (Compare (1) and (2) in the Table~\ref{tab:dpr_init}). We highlight the performance improvements for both answer generation accuracy and retrieval recall scores, where the Covid-19 domain has the largest improvements. We also compare the finetuning \textit{RAG-end2end} model with the RAG-original model, which was first initialized with the domain-adapted DPR models  (Compare (2) and (3) in Table~\ref{tab:dpr_init}). This comparison shows that \textit{RAG-end2end} training mechanism can outperform the RAG-original mechanism that uses a domain-adapted DPR. The results further highlight the importance of \textit{RAG-end2end} mechanism in domain adaptation where we do not need to train the DPR model separately.

\section{Discussion}
\subsection{Role of retriever in domain adaptation}
As the results suggest, the retriever plays an essential role in domain adaptation for open-domain QA. It is clear that \textit{RAG-end2end} training improves the results since it can update the embeddings and the indexing of the knowledge base. Compared with the original RAG finetuning, \textit{RAG-end2end} improves the performance in all datasets. The main reason for this could be that neural retrievers such as DPR, which are trained on public datasets, struggle to perform well on domain-specific datasets. Our results also highlight an important aspect related to the performance of the stand-alone DPR for document retrieval. It shows that \textit{RAG-end2end} can improve the domain adaptation of DPR better that finetuning the DPR on its own mechanism.

\subsection{Cost of end2end retriever adaptation}
It is important to note that \textit{RAG-end2end} fine tuning can be expensive if the number of passages in the external knowledge base is large. If there are millions of passages, it would be beneficial to have a dedicated number of GPUs that complete the re-encoding process. Re-indexing with the FAISS library also depends on the number of cores in the CPUs. When having access to strong enough computational power, it is better to use \textit{RAG-end2end} since we can directly use passages in a knowledge base and question-answer pairs to train both the retriever and the reader. Then we also do not need to generate synthetic question-answer pairs related to passages that are required to train the DPR.

Although the  RETRO~\cite{borgeaud2021improving} authors claim that frozen BERT embedding is sufficient for retrieval augmented models, our results suggest that for domain-specific models to perform well, a domain-adapted retriever component is beneficial. In future work, it is important to explore how the models like RETRO~\cite{borgeaud2021improving} perform on domain-specific scenarios going beyond general-purpose datasets.

\subsection{Comparing the RAG-original with RAG-end2end on an In-domain dataset}

\begin{table}[h]
\vspace{-10pt}
\centering
\def\arraystretch{1.3}
\resizebox{\columnwidth}{!}{%
\begin{tabular}{lllll}
\multicolumn{1}{c|}{Model Name}& \multicolumn{1}{c|}{EM}    & \multicolumn{1}{c|}{F1}    & \multicolumn{1}{c|}{Top-5} & \multicolumn{1}{c}{Top-20} \\ \hline
\multicolumn{5}{c}{SQUAD Open-Domain}                                                                                                                                                                \\ \hline
\multicolumn{1}{l|}{(1) RAG-original}                                          & \multicolumn{1}{l|}{28.12}  & \multicolumn{1}{l|}{39.42}  & \multicolumn{1}{l|}{59.64}  & \multicolumn{1}{l}{72.38}   \\

\multicolumn{1}{l|}{(2) RAG-end2end}                                      & \multicolumn{1}{l|}{40.02}  & \multicolumn{1}{l|}{52.63} & \multicolumn{1}{l|}{75.79} & \multicolumn{1}{l}{85.57}  
\end{tabular} }%
\caption{Open-Domain performance comparison between RAG-original and RAG-end2end. We used SQAD dataset in ODQA manner to conduct our experiments.}
\label{tab:open-domain}
\vspace{-10pt}
\end{table}

Although our work is mainly focused on the domain adaptation of RAG for specific domains, we also explored whether the end2end training would improve the overall results of an in-domain dataset. Since the original RAG model uses a DPR model that is trained on a Wikipedia-based Natural Questions dataset, we consider this in-domain. Although SQUAD~\cite{rajpurkar2016squad} dataset is a machine comprehension dataset, we adapted the SQUAD dataset to perform ODQA. First, we extracted the contexts related to each question-answer pair and created an external knowledge base. Then we split the knowledge base into 100-words passages. Our final knowledge base consists of 30K passages. As illustrated in Table~\ref{tab:open-domain}, we compared the performance of  RAG-original and \textit{RAG-end2end} on the tasks of answer generation and retrieving correct documents. As the results suggested, \textit{RAG-end2end} performs better than RAG-original even in other Wikipedia-based datasets. This could be due to \textit{RAG-end2end} updating the context encoder and embeddings during the training process.

\section{Conclusion and Future Work}
In this paper, we proposed a  novel extension of RAG: \textit{RAG-end2end}, which, unlike RAG, does joint training of the retriever and
generator for the end QA task and domain
adaptation. We showed that the \textit{RAG-end2end} could improve DPR performance better than fine-tuning the DPR independently. This allows for the training of DPR models with QA pairs and eliminates the need for gold-standard passages related to questions. We also highlighted that the addition of a re-construction auxiliary signal further improves both the retriever and the final answer generation accuracies. We evaluate our approach with three datasets from different domains (COVID-19, News, and Conversations), showing that \textit{RAG-end2end} achieves significant performance improvements in all three domains compared to the original RAG implementation. In addition, we conducted several other experiments to validate our approach comprehensively. Overall, our results show that our approach is stable and generalizable across different domains. Our experiments highlight the importance of the RAG's retriever component in domain-specific question answering.

Based on our findings, we suggest three directions for future research in domain adaptation of RAG Models.
Firstly, 
we consider it worthwhile to explore \textit{RAG-end2end} on other tasks like Fact Checking~\cite{lewis2020retrieval}, Summarisation~\cite{shuster2021retrieval}, and conversational response generation~\cite{xu2021beyond} where the original RAG has  shown interesting results. Secondly, it is important to explore generative capabilities with qualitative metrics. This could be aligned with research areas like measuring factual consistency~\cite{kryscinski2019evaluating,cao2022hallucinated} and hallucinations~\cite{cao2022hallucinated,shuster2021retrieval,nie2019simple} of generative language models. Future work could explore whether updating the retriever and document embeddings during the training phase could improve factual consistency and reduce hallucinations in final generations. Thirdly, the improvement of RAG with our extension (\textit{RAG-end2end}) highlights the importance of the retriever in the RAG architecture, which motivates us to improve the retriever part further in future work. Also, as the statement re-construction signal acts as a good auxiliary signal, we encourage exploring other auxiliary signals, which could improve the overall performance of RAG models.

\balance{}
\bibliography{anthology,tacl2021}
\bibliographystyle{acl_natbib}
\balance

\newpage

\onecolumn
\appendix
\section{Appendix}
\begin{figure}[h!]
\centering
\includegraphics[width=1.0\textwidth]{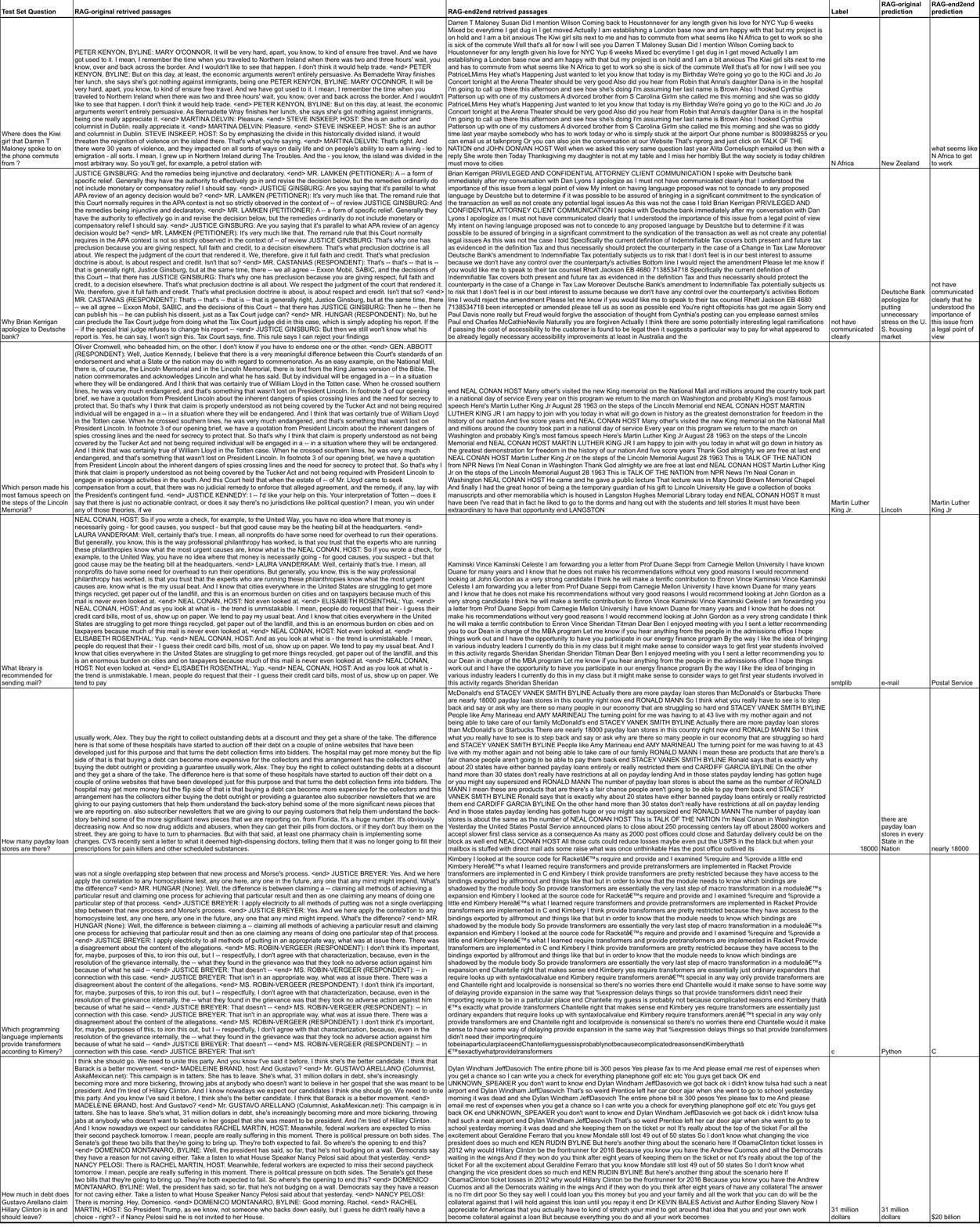}
\vspace{-25pt}
\caption{Predicted answers and retrieved passages for a set of questions from the conversational domain~\cite{wu2021qaconv}.}
\label{fig:examples}
\end{figure}

\end{document}